\documentclass{article}

\usepackage{nips15submit_e}

\usepackage[utf8]{inputenc} 
\usepackage[T1]{fontenc}    
\usepackage{hyperref}       
\usepackage{url}            
\usepackage{booktabs}       
\usepackage{amsfonts}       
\usepackage{nicefrac}       
\usepackage{microtype}      
\usepackage{graphicx}

\usepackage{amsmath,amssymb,amsthm,mathrsfs,amsfonts,dsfont} 
\newcommand{\tabcaption}{\def\@captype{table}\caption}
\newcommand{\figcaption}{\def\@captype{figure}\caption}
\newcommand{\mr}{\mathrm}
\newcommand{\mb}{\mathbf}
\newcommand{\mc}{\mathcal}
\newcommand{\md}{\mathds}

\newcommand{\bs}{\boldsymbol}

\title{Functional Hashing for Compressing \\ Neural Networks }

\author{
  Lei Shi, Shikun Feng, Zhifan Zhu\\
  Baidu, Inc.\\
  \texttt{\{shilei06,\ fengshikun,\ zhuzhifan\}@baidu.com} \\
}

\begin{document}

\nipsfinalcopy 

\maketitle

\begin{abstract}
As the complexity of deep neural networks (DNNs) trend to grow to absorb the increasing sizes of data, memory and energy consumption has been receiving more and more attentions for industrial applications, especially on mobile devices. This paper presents a novel structure based on functional hashing to compress DNNs, namely FunHashNN. For each entry in a deep net, FunHashNN uses multiple low-cost hash functions to fetch values in the compression space, and then employs a small reconstruction network to recover that entry. The reconstruction network is plugged into the whole network and trained jointly. FunHashNN includes the recently proposed HashedNets \cite{HashNet_ICML2015} as a degenerated case, and benefits from larger value capacity and less reconstruction loss. We further discuss extensions with dual space hashing and multi-hops. On several benchmark datasets, FunHashNN demonstrates high compression ratios with little loss on prediction accuracy.
\end{abstract}

\section{Introduction}\label{sec:introduction}
Deep Neural networks (DNNs) have been receiving ubiquitous success in wide applications, ranging from computer vision \cite{DNNonCV_NIPS2012}, to speech recognition \cite{DNNonSpeech_IEEE_2012}, natural language processing \cite{DNNonNLP_JMLR2011}, and domain adaptation \cite{DNNonDomain_ICML2011}.  As the sizes of data mount up, people usually have to increase the number of parameters in DNNs so as to absorb the vast volume of supervision. High performance computing techniques are investigated to speed up DNN training, concerning optimization algorithms, parallel synchronisations on clusters w/o GPUs, and stochastic binarization/ternarization, etc \cite{BinaryConnect_NIPS2015,8BitPara_ICLR2016,TenaryConnect_ICLR2016}.

On the other hand the memory and energy consumption is usually, if not always, constrained in industrial applications \cite{Mobile_ICLR2016,DeepFried_ICCV2015}. For instance, for commercial search engines (e.g., Google and Baidu) and recommendation systems (e.g., NetFlix and YouTube), the ratio between the increased model size and the improved performance should be considered given limited online resources. Compressing the model size becomes more important for applications on mobile and embedded devices \cite{SongHan_ICLR2016,Mobile_ICLR2016}. Having DNNs running on mobile apps owns many great features such as better privacy, less network bandwidth and real time processing. However, the energy consumption of battery-constrained mobile devices is usually dominated by memory access, which would be greatly saved if a DNN model can fit in on-chip storage rather than DRAM storage (c.f. \cite{SongHan_ICLR2016,SongHan_NIPS2015} for details).

A recent trend of studies are thus motivated to focus on compressing the size of DNNs while mostly keeping their predictive performance \cite{SongHan_ICLR2016,Mobile_ICLR2016,DeepFried_ICCV2015}. With different intuitions, there are mainly two types of DNN compression methods, which could be used in conjunction for better parameter savings. The first type tries to revise the training target into more informative supervision using \emph{dark knowledge}. In specific, Hinton \emph{et al.} \cite{Distll_NIPS2014} suggested to train a large network ahead, and distill a much smaller model on a combination of the original labels and the soft-output by the large net. The second type observes the redundancy existence in network weights \cite{HashNet_ICML2015,RedundDNN_NIPS2013}, and exploits techniques to constrain or reduce the number of free-parameters in DNNs during learning. This paper focuses on the latter type.

To constrain the network redundancy, efforts \cite{RedundDNN_NIPS2013,FastFood_ICML2013,DeepFried_ICCV2015} formulated an original weight matrix into either low-rank or fast-food decompositions. Moreover \cite{SongHan_ICLR2016,SongHan_NIPS2015} proposed a simple-yet-effective pruning-retraining iteration during training, followed by quantization and fine-tuning. Chen \emph{et al.} \cite{HashNet_ICML2015}  proposed HashedNets to efficiently implement parameter sharing prior to learning, and showed notable compression with much less loss of accuracy than low-rank decomposition. More precisely, prior to training, a hash function is used to randomly group (virtual) weights into a small number of buckets, so that all weights mapped into one hash bucket directly share a same value. HashedNets was further deliberated in frequency domain for compressing convolutional neural networks in \cite{HashNet_NIPS2015}. 

In applications, we observe HashedNets compresses model sizes greatly at marginal loss of accuracy for some situations, whereas also significantly loses accuracy for others. After revisiting its mechanism, we conjecture this instability comes from at least three factors. First, hashing and training are disjoint in a two-phase manner, i.e., once inappropriate collisions exist, there may be no much optimization room left for training. Second, \emph{one single hash function} is used to fetch a single value in the compression space, whose collision risk is larger than multiple hashes \cite{MultiHash_6}. Third, parameter sharing within a buckets implicitly uses \emph{identity mapping} from the hashed value to the virtual entry.

This paper proposes an approach to relieve this instability, still in a two-phase style for preserving efficiency. Specifically, we use \emph{multiple hash functions} \cite{MultiHash_6} to map per virtual entry into multiple values in compression space. Then an additional network plays in a \emph{mapping function} role from these hashed values to the virtual entry before hashing, which can be also regarded as ``reconstructing'' the virtual entry from its multiple hashed values. Plugged into and jointly trained within the original network, the reconstruction network is of a comparably ignorable size, i.e., at low memory cost. 

This functional hashing structure includes HashedNets as a degenerated special case, and facilitates less value collisions and better value reconstruction.  Shortly denoted as FunHashNN, our approach could be further extended with dual space hashing and multi-hops. Since it imposes no restriction on other network design choices (e.g. dropout and weight sparsification), FunHashNN can be considered as a standard tool for DNN compression. Experiments on several datasets demonstrate promisingly larger reduction of model sizes and/or less loss on prediction accuracy, compared with HashedNets.

\section{Background}\label{sec:background}
\vspace{-2mm}
\paragraph{Notations.} Throughout this paper we express scalars in regular ($A$ or $b$), vectors in bold ($\mb{x}$), and matrices in capital bold ($\mb{X}$). Furthermore, we use $x_i$ to represent the $i$-th dimension of vector $\mb{x}$, and use $X_{ij}$ to represent the $(i,j)$-th entry of matrix $\mb{X}$. Occasionally, $[\mb{x}]_i$ is also used to represent the $i$-th dimension of vector $\mb{x}$ for specification clarity . Notation $\md{E}[\cdot]$ stands for the expectation operator.  

\paragraph{Feed Forward Neural Networks.} We define the forward propagation of the $\ell$-th layer as
\vspace{-1mm}\begin{eqnarray}\label{eq:FFNN}
 a_i^{\ell+1} = f(z_i^{\ell+1}), \quad\mr{with}\quad z_i^{\ell+1}=b_i^{\ell+1} + \sum_{j=1}^{d^{\ell}} V_{ij}^{\ell} a_j^{\ell}, \quad \mr{for} \ \forall i\in[1,d^{\ell+1}].
\end{eqnarray}
For each $\ell$-th layer, $d^{\ell}$ is the output dimensionality, $\mb{b}^{\ell}$ is the bias vector, and $\mb{V}^{\ell}$ is the (\emph{virtual}) weight matrix in the $\ell$-th layer. Vectors $\mb{z}^{\ell},\mb{a}^{\ell}\in\md{R}^{d^{\ell}}$ denote the units before and after the activation function $f(\cdot)$. Typical choices of $f(\cdot)$ include rectified linear unit (ReLU) \cite{ReLU_ICML2010}, sigmoid and tanh \cite{NNPR_1995}.

\paragraph{Feature Hashing} has been studied as a dimension reduction method for reducing model storage size without maintaining the mapping matrices like random projection \cite{Hash_Shi_JMLR2009,Hash_Wein_ICML2009}. Briefly, it maps an input vector $\mb{x}\in\md{R}^n$ to a much smaller feature space via $\bs{\phi}:\md{R}^n \rightarrow \md{R}^K$ with $K\ll n$. Following the definition in \cite{Hash_Wein_ICML2009}, the mapping $\bs{\phi}$ is a composite of two approximate uniform hash functions $h:\md{N}\rightarrow\{1,\ldots,K\}$ and $\xi:\md{N}\rightarrow\{-1,+1\}$. The $j$-th element of $\bs{\phi}(\mb{x})$ is defined as:
\vspace{-1mm}\begin{eqnarray}\label{eq:OriginalFeatureHash}
 \left[\bs{\phi}(\mb{x})\right]_j = \sum_{i:h(i)=j} \xi(i)x_i.
\end{eqnarray}
As shown in \cite{Hash_Wein_ICML2009}, a key property is its inner product preservation, which we quote and restate below.

\textbf{\emph{Lemma [Inner Product Preservation of Original Feature Hashing]}} With the hash defined by Eq.~(\ref{eq:OriginalFeatureHash}), the hash kernel is unbiased, i.e., $\md{E}_{\bs{\phi}}[\bs{\phi}(\mb{x})^{\top}\bs{\phi}(\mb{y})]=\mb{x}^{\top}\mb{y}$. Moreover, the variance is $\mr{var}_{\mb{x},\mb{y}}=\frac{1}{K}\sum_{i\neq j}\left(x_i^2y_j^2+x_iy_ix_jy_j\right)$, and thus $\mr{var}_{\mb{x},\mb{y}}=\mc{O}(\frac{1}{K})$ if $||\mb{x}||_2=||\mb{y}||_2=\mr{const}$.

\paragraph{HashedNets in \cite{HashNet_ICML2015}.} As illustrated in Figure~\ref{fig:model}(a), HashedNets randomly maps network weights into a smaller number of groups prior to learning, and the weights in a same group share a same value thereafter. A naive implementation could be trivially achieved by maintaining a secondary matrix that records the group assignment, at the expense of additional memory cost however. HashedNets instead adopts a hash function that requires no storage cost with the model. Assume there is a finite memory budge $K^{\ell}$ per layer to represent $\mb{V}^{\ell}$, with $K^{\ell}\ll (d^{\ell}+1)d^{\ell+1}$. We only need to store a weight vector $\mb{w}^{\ell}\in\md{R}^{K^{\ell}}$, and assign $V_{ij}^{\ell}$ an element in $\mb{w}^{\ell}$ indexed by a hash function $h^{\ell}(i,j)$, namely
\vspace{-1mm}\begin{eqnarray}\label{eq:HashedNets}
 V_{ij}^{\ell} = \xi^{\ell}(i,j)\cdot w_{h^{\ell}(i,j)}^{\ell},
\end{eqnarray}
where hash function $h^{\ell}(i,j)$ outputs an integer within $[1,K^{\ell}]$. Another independent hash function $\xi^{\ell}(i,j):(d^{\ell+1}\times d^{\ell})\rightarrow\pm1$ outputs a sign factor, aiming to reduce the bias due to hash collisions \cite{Hash_Wein_ICML2009}. The resulting matrix $\mb{V}^{\ell}$ is \emph{virtual}, since $d^{\ell}$ could be increased without increasing the \emph{actual} number of parameters in $\mb{w}^{\ell}$ once the compression space size $K^{\ell}$ is determined and fixed. 

Substituting Eq.~(\ref{eq:HashedNets}) into Eq.~(\ref{eq:FFNN}), we have $z_i^{\ell+1}=b_i^{\ell+1}+\sum_{j=1}^{d^{\ell}}\xi^{\ell}(i,j)w^{\ell}_{h^{\ell}(i,j)}a_j^{\ell}$. During training, $\mb{w}^{\ell}$ is updated by back propagating the gradient via $\mb{z}^{\ell+1}$ (and the virtual $\mb{V}^{\ell}$). Besides, the activation function $f(\cdot)$ in Eq.~(\ref{eq:FFNN}) was kept as ReLU in \cite{HashNet_ICML2015} to further relieve the hash collision effect through a sparse feature space. In both \cite{HashNet_ICML2015} and this paper, the open source \emph{xxHash}\footnote{http://cyan4973.github.io/xxHash/} is adopted as an approximately uniform hash implementation with low cost. 

\begin{minipage}[!h]{\textwidth}
\centering
 \begin{minipage}{.95\textwidth}
 \centering
   \includegraphics[width=.9\textwidth]{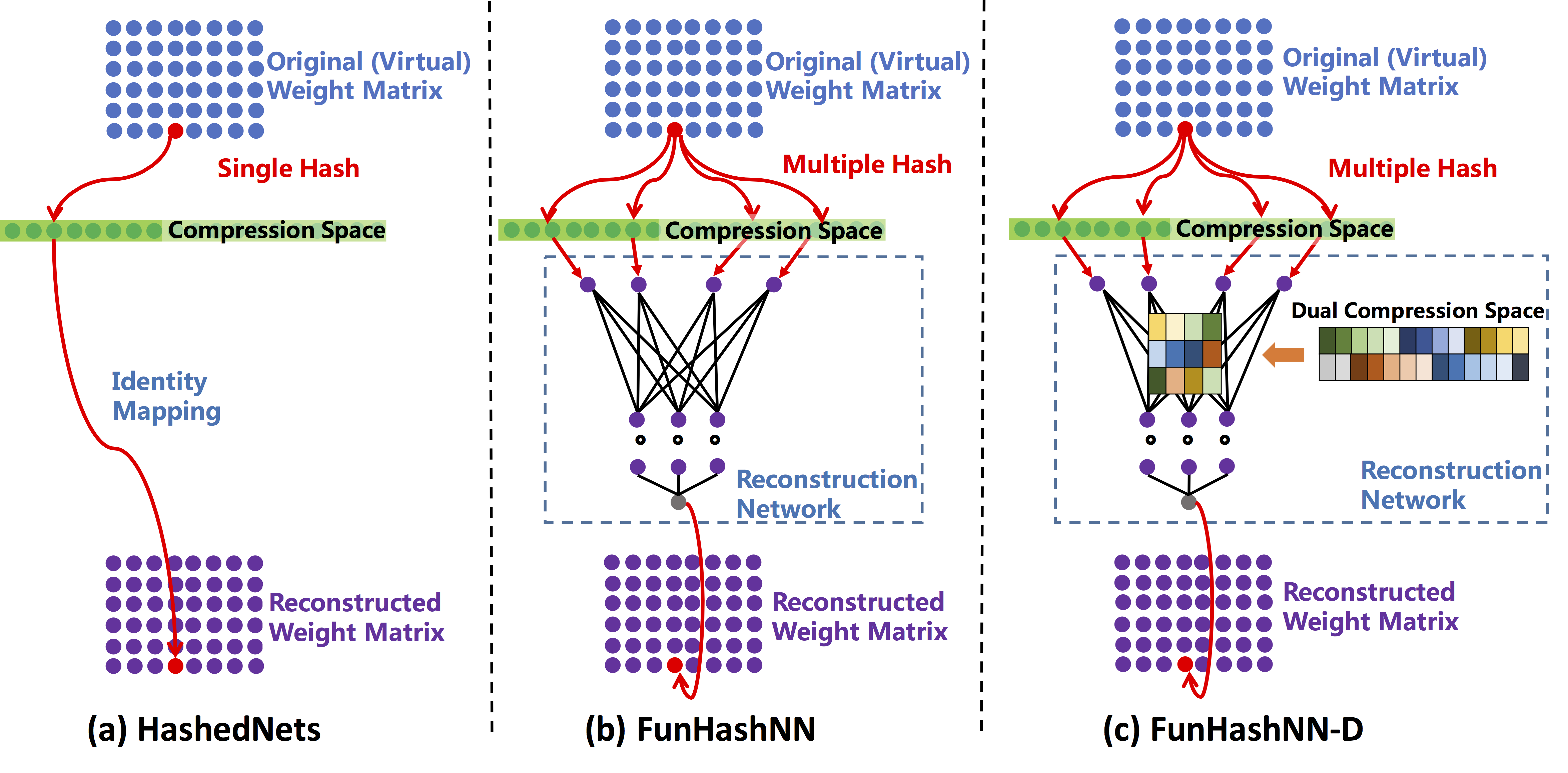}
   \vspace{-4mm}
 \makeatletter\def\@captype{figure}\makeatother
 \caption{Illustrations of hashing approaches for neural networks compression. (a) HashedNets \cite{HashNet_ICML2015}. (b) our FunHashNN. (c) our FunHashNN with dual space hashing. (Best viewed in color)} \label{fig:model}
 \end{minipage}
\end{minipage}

\section{Functional Hashing for Neural Network Compression}\label{sec:formulation}
\vspace{-2mm}
\subsection{Structure Formulation}
\vspace{-1mm}
For clarity, we will focus on a single layer throughout and drop the super-script $\ell$.  Still, vector $\mb{w}\in\md{R}^K$ denotes parameters in the compression space. The key difference between FunHashNN and HashedNets \cite{HashNet_ICML2015} lies in (i) how to employ hash functions, and (ii) how to map from $\mb{w}$ to $\mb{V}$:
\begin{itemize}
 \item Instead of adopting one pair of hash function $(h,\xi)$ in Eq.~(\ref{eq:HashedNets}), we use a set of multiple pairs of independent random hash functions. Let's say there are $U$ pairs of mappings $\{h_u, \xi_u\}_{u=1}^U$, each $h_u(i,j)$ outputs an integer within $[1,K]$, and each $\xi_u(i,j)$ selects a sign factor.
 \item Eq.~(\ref{eq:HashedNets}) of HashedNets employs an identity mapping between one element in $\mb{V}$ and one hashed value, i.e., $V_{ij}=\xi(i,j)w_{h(i,j)}$. In contrast, we use a multivariate function $g(\cdot)$ to describe the mapping from multiple hashed values $\{\xi_u(i,j)w_{h_u(i,j)}\}_{u=1}^U$ to $V_{ij}$. Specifically,
 	\begin{eqnarray}\label{eq:FunHashNN}
	 V_{ij} = g\left(\left[\xi_1(i,j)w_{h_1(i,j)},\ \ldots,\ \xi_U(i,j)w_{h_U(i,j)}\right];\ \bs{\alpha}\right).
	\end{eqnarray}
Therein, $\bs{\alpha}$ is referred to as the parameters in $g(\cdot)$. Note that the input $\xi_u(i,j)w_{h_u(i,j)}$ is order sensitive from $u=1$ to $U$. We choose $g(\cdot)$ to be a multi-layer feed forward neural network, and other multivariate functions may be considered as alternatives.
\end{itemize}

As a whole, Figure~\ref{fig:model}(b) illustrates our FunHashNN structure, which can be easily plugged in any matrices of DNNs. Note that $\bs{\alpha}$ in the reconstruction network $g(\cdot)$ is of a much smaller size compared to $\mb{w}$. For instance, a setting with $U=4$ and a 1-layer $g(\cdot;\bs{\alpha})$ of $\bs{\alpha}\in\mc{R}^4$ performs already well enough in experiments. In other words, Eq.~(\ref{eq:FunHashNN}) just uses an ignorable amount of additional memory to describe a functional $\mb{w}$-to-$\mb{V}$ mapping, whose properties will be further explained in the sequel.

\subsection{Training Procedure}
\vspace{-1mm}
The parameters in need of updating include $\mb{w}$ in the compression and $\bs{\alpha}$ in $g(\cdot)$. Training FunHashNN is equivalent to training a standard neural network, except that we need to forward/backward-propagate values related to $\mb{w}$ through $g(\cdot)$ and the virtual matrix $\mb{V}$. 

\paragraph{Forward Propagation.} Substituting Eq.~(\ref{eq:FunHashNN}) into Eq.~(\ref{eq:FFNN}), we still omit the super-script $\ell$ and get
\vspace{-1mm}\begin{eqnarray}
 z_i = b_i + \sum_{j=1}^d a_jV_{ij} =  b_i + \sum_{j=1}^d a_j \cdot g\left(\left[\xi_1(i,j)w_{h_1(i,j)},\ \ldots,\ \xi_U(i,j)w_{h_U(i,j)}\right];\ \bs{\alpha}\right).
\end{eqnarray}

\paragraph{Backward Propagation.} Denote $\mc{L}$ as the final loss function, e.g., cross entropy or squared loss, and suppose $\delta_i=\frac{\partial \mc{L}}{\partial z_i}$ is already available by back-propagation from layers above. The derivatives of $\mc{L}$ with respect to $\mb{w}$ and $\bs{\alpha}$ are computed by
\vspace{-1mm}\begin{eqnarray}
 \frac{\partial \mc{L}}{\partial \mb{w}} = \sum_i\sum_j a_j\delta_i\frac{\partial V_{ij}}{\partial \mb{w}},
 \qquad
 \frac{\partial \mc{L}}{\partial \bs{\alpha}} = \sum_i\sum_j a_j\delta_i\frac{\partial V_{ij}}{\partial \bs{\alpha}},
\end{eqnarray}
where, since we choose $g(\cdot)$ as a multilayer neural network, derivatives $\frac{\partial V_{ij}}{\partial \mb{w}}$ and $\frac{\partial V_{ij}}{\partial \bs{\alpha}}$ can be calculated through the small network $g(\cdot)$ in a standard back-propagation manner.

\paragraph{Complexity.} Concerning time and memory cost, FunHashNN roughly has the same complexity as HashedNets, since the small network $g(\cdot)$ is quite light-weighted. One key variable factor is the way to implement multiple hash functions. On one hand, if they are calculated online, then FunHashNN requires little additional time if tackling them in parallel. On the other, if they are pre-computed and stored in dicts to avoid hashing time cost, the multiple hash functions of FunHashNN demand more storage space. In application, we suggest to pre-compute hashes during offline training for speedup, and to compute hashes in parallel during online prediction for saving memory under limited budget.

\subsection{Property Analysis}
\vspace{-1mm}
In this part, we try to depict the properties of our FunHashNN from several aspects to help understanding it, especially in comparison with HashedNets \cite{HashNet_ICML2015}.

\paragraph{Value Collision.} It should be noted, both HashedNets and FunHashNN conduct hashing prior to training, i.e., in a two-phase manner. Consequently, it would be unsatisfactory if hashing collisions happen among important values. For instance in natural language processing tasks, one may observe wired results if there are many hashing collisions among embeddings (which form a matrix) of frequent words, especially when they are not related at all. In the literature, multiple hash functions are known to perform better than one single function \cite{MultiHash_3,MultiHash_6,MultiHash_INFOCOM2001}. In intuition, when we have multiple hash functions, the items colliding in one function are hashed
differently by other hash functions.

\paragraph{Value Reconstruction.} In both HashedNets and FunHashNN, the hashing trick can be viewed as a reconstruction of the original parameter $\mb{V}$ from $\mb{w}\in\mc{R}^{K}$. In this sense, the approach with a lower reconstruction error is preferred\footnote{One might argue that there exists redundancy in $\mb{V}$, whereas here we could imagine $\mb{V}$ is already structured and filled by values with least redundancy.}. Then we have at least the following two observations:
\begin{itemize}
 \item \textbf{The maximum number of possible distinct values} output by hashing intuitively explains the modelling capability \cite{Hash_Shi_JMLR2009}. For HashedNets, considering the sign hashing function $\xi(\cdot)$, we have at most $2K$ possible distinct values of Eq.~(\ref{eq:HashedNets}) to represent elements in $\mb{V}$. In contrast, since there are multiple ordered hashed inputs, FunHashNN has at most $(2K)^U$ possible distinct values of Eq.~(\ref{eq:FunHashNN}). Note that the memory size $K$ is the same for both. 
 
 \item \textbf{The reconstruction error} may be difficult to analyzed directly, since the hashing mechanism is trained jointly within the whole network. However, we observe $g\left(\left[\xi_1(i,j)w_{h_1(i,j)},\ \ldots,\ \xi_U(i,j)w_{h_U(i,j)}\right];\ \bs{\alpha}\right)$ degenerates to $g(\xi_1(i,j)w_{h_1(i,j)})$ if we assign zeros to all entries in $\bs{\alpha}$ unrelated to the 1st input dimension. Since $g(\xi_1(i,j)w_{h_1(i,j)})$ depends only on one single pair of hash functions, it is conceptually equivalent to HashedNets.   Consequently, including HashedNets as a special case, FunHashNN with freely adjustable $\bs{\alpha}$ is able to reach a lower reconstruction error to fit the final accuracy better.
\end{itemize}

\paragraph{Feature Hashing.} In line with previous work \cite{Hash_Shi_JMLR2009,Hash_Wein_ICML2009}, we compare HashedNets and FunHashNN in terms of feature hashing. For specification clarity, we drop the sign hashing functions $\xi(\cdot)$ below for both methods, the analysis with which is straightforward by replacing $K$ hereafter with $2K$. 
\begin{itemize}
 \item For HashedNets, one first defines a hash mapping function $\bs{\phi}_i^{(1)}(\mb{a})$, whose $k$-th element is 
 	\vspace{-1mm}\begin{eqnarray}\label{eq:featurehash_hashnet1}
	 \left[\bs{\phi}_i^{(1)}(\mb{a})\right]_k \triangleq \sum_{j:h(i,j)=k} a_j, \quad \mr{for} \quad k=1,\ldots,K.
	\end{eqnarray}
	Thus $z_i$ by HashedNets can be computed as the inner product (details c.f. Section~4.3 in \cite{HashNet_ICML2015})
	\vspace{-1mm}\begin{eqnarray}\label{eq:featurehash_hashnet2}
	 z_i = \mb{w}^{\top} \bs{\phi}_i^{(1)}(\mb{a}).
	\end{eqnarray}
 \item For FunHashNN, we first define a hash mapping function $\bs{\phi}_i^{(2)}(\mb{a})$. Different from a $K$-dim output in Eq.~(\ref{eq:featurehash_hashnet1}), it is of a much larger size $K^U$,  with $\left(\sum_{u=1}^U k_uK^{(u-1)}\right)$-th element as
 	\vspace{-1mm}\begin{eqnarray}\label{eq:featurehash_funhashnn1}
	 \left[\bs{\phi}_i^{(2)}(\mb{a})\right]_{\sum_{u=1}^U k_uK^{(u-1)}}  \triangleq \sum_{\substack{j:h_1(i,j)=k_1 \\ \ \ h_2(i,j)=k_2 \\ \ \ \ldots \\ \ \ h_U(i,j)=k_U}} a_j, \quad \mr{for}\quad \forall u, \ k_u=1,\ldots,K.
	\end{eqnarray}
	Second, we define vector $\mb{g}_{\bs{\alpha}}(\mb{w})$ still of length $K^U$, whose $\left(\sum_{u=1}^U k_uK^{(u-1)}\right)$-th entry is 
	\vspace{-1mm}\begin{eqnarray}\label{eq:featurehash_funhashnn2}
	 \left[\mb{g}_{\bs{\alpha}}(\mb{w})\right]_{\sum_{u=1}^U k_uK^{(u-1)}}  \triangleq g\left(w_{k_1},w_{k_2},\ldots,w_{k_U}; \ \bs{\alpha}\right), \quad \mr{for}\quad \forall u, \ k_u=1,\ldots,K.
	\end{eqnarray}
	Thus $z_i$ by FunHashNN can be computed as the following inner product
	\vspace{-1mm}\begin{eqnarray}\label{eq:featurehash_funhashnn3}
	 z_i = {\mb{g}_{\bs{\alpha}}(\mb{w})}^{\top} \bs{\phi}_i^{(2)}(\mb{a}).
	\end{eqnarray}
	The difference between Eq.~(\ref{eq:featurehash_hashnet2}) and Eq.~(\ref{eq:featurehash_funhashnn3}) further explains the above discussion about ``the maximum number of possible distinct values''.  
\end{itemize}

\subsection{Extensions}
\vspace{-2mm}
\paragraph{Hashing on Dual Space.} If considering a linear model $f(\mb{x};\bs{\theta})=\bs{\theta}^{\top}\mb{x}$, one can not only deliver analysis like Bayesian or hashing on input feature space of $\mb{x}$, but also do similarly on the \emph{dual space} of $\bs{\theta}$ \cite{PRML_2006}. We now revisit the ``reconstruction'' network $g(\mb{x}_{ij}; \bs{\alpha})$ in Eq.~(\ref{eq:FunHashNN}), where vector $\mb{x}_{ij}$ concatenates the hashed values $\xi_u(i,j)w_{h_u(i,j)}$ for $u=1,\ldots,U$. What we did in Eq.~(\ref{eq:FunHashNN}) is in fact hashing $(i,j)$ through $\mb{w}$ to get the input feature of $g(\cdot)$. In analogy, we can also hash $(i,j)$ to fetch parameters of $g(\cdot)$, namely we have a new ``reconstruction'' network in the following form:
\begin{eqnarray}
 V_{ij} = g(\mb{x}_{ij};\bs{\alpha}_{ij}), \quad\mr{with}\quad [\mb{x}_{ij}]_u = \xi_u(i,j)w_{h_u(i,j)}\quad \mr{and} 
 	\quad [\bs{\alpha}_{ij}]_r = \xi'_r(i,j) w'_{h'_r(i,j)},
\end{eqnarray}
where $\{\xi'_r(\cdot), h'_r(\cdot)\}$ are additional multiple pairs of hash functions applied on $\bs{\alpha}$, and $\mb{w}'$ is an additional vector in the compression space of $\bs{\alpha}$. The size of $\bs{\alpha}_{ij}$ remains the same as previous. Using this trick, the maximum number of possible distinct values of $\mb{V}$ further increases exponentially, so that FunHashNN has more potential ability to fit the prediction well. We denote FunHashNN with dual space hashing shortly as FunHashNN-D, and illustrate its structure in Figure~\ref{fig:model}(c).

\paragraph{Multi-hops.} We conjecture that FunHashNN could be used in a multi-hops structure, by imagining $\mb{w}$ in the compression space plays a \emph{virtual} role similar to $\mb{V}$. Specifically, we can build another level of hash functions $\left\{\xi^{(1)}_u(\cdot),h^{(1)}_u(\cdot)\right\}$ and compression space $\mb{w}^{(1)}$. Thereafter, each entry in $\mb{w}$ is hashed into multiple values in $\mb{w}^{(1)}$ via $\left\{\xi^{(1)}_u(\cdot),h^{(1)}_u(\cdot)\right\}$. Then another reconstruction network $g^{(1)}(\cdot)$ is used to learn the mapping from the hashed values in $\mb{w}^{(1)}$ to the corresponding entry in $\mb{w}$. 

This procedure can be implemented recursively. If there are in total $M$-hops, what we need to save in fact just includes a (possibly much more smaller) vector $\mb{w}^{(M)}$ at the final hop, a series of $M$ small reconstruction networks $\{g^{(m)}(\cdot)\}_{m=1}^M$, and a series of hashing functions. In contrast, the multi-hops version of HashedNets is equivalent to just adjusting the compression ratio, or say the size $K$.

\vspace{-2mm}
\section{Related Work}\label{sec:relatedwork}
\vspace{-3mm}
Recent studies have confirmed the redundancy existence in the parameters of deep neural networks. Denil \emph{et al.} \cite{RedundDNN_NIPS2013} decomposed a matrix in a fully-connected layers as the product of two low-rank matrices, so that the number of parameters decreases linearly as the latent dimensionality decreases. More structured decompositions Fastfood \cite{FastFood_ICML2013} and Deep Fried \cite{DeepFried_ICCV2015} were proposed not only to reduce the number of parameters, but also to speed up matrix multiplications. More recently, Han \emph{et al.} \cite{SongHan_ICLR2016,SongHan_NIPS2015} proposed to iterate pruning-retraining during training DNNs, and used quantization and fine-tuning as a post-processing step. Huffman coding and hardware implementation were also considered. In order to mostly keep accuracy, the authors suggested multiple rounds of pruning-retraining. That is, for little accuracy loss, we have to prune slowly enough and thus suffer from increased training time. Again, the most related work to ours is HashedNets \cite{HashNet_ICML2015}, which was then extended in \cite{HashNet_NIPS2015} to random hashing in frequency domain for compressing convolutional neural networks. Either HashedNets or FunHashNN could be combined in conjunction with other techniques for better compression.

Extensive studies have been made on constructing and analyzing multiple hash functions, which 
have shown better performances over one single
hash function \cite{MultiHash_6}. One multi-hashing algorithm, $d$-random
scheme \cite{MultiHash_3}, uses only one hash table but $d$ hash functions, pretty similar
to our settings. One choice alternative to $d$-random is the $d$-left algorithm proposed in 
\cite{MultiHash_INFOCOM2001}, used for improving IP lookups. Hashing algorithms for natural
language processing are also studied in \cite{Sketch_NLP_EMNLP2012}. Papers \cite{Hash_Shi_JMLR2009,Hash_Wein_ICML2009} investigated feature hashing (a.k.a. the hashing trick), providing useful bounds and feasible results.

\vspace{-2mm}
\section{Experiments}\label{sec:experiments}
\vspace{-3mm}
We conduct extensive experiments to evaluate FunHashNN on DNN compression. Codes for fully reproducibility will be open source soon after necessary polishment.

\subsection{Environment Descriptions}
\vspace{-2mm}
\paragraph{Datasets.} Three benchmark datasets \cite{DataSet_ICML2007} are considered here, including (1) the original \texttt{MNIST} hand-written digit dataset, (2) dataset \texttt{BG-IMG} as a variant to \texttt{MNIST}, and (3) binary image classification dataset \texttt{CONVEX}. For all datasets, we use prespecified training and testing splits. In particular, the original \texttt{MNIST} dataset has \#train=60,000 and \#test=10,000, while the remaining both have \#train=12,000 and \#test=50,000. Moreover, collected from a commercial search engine, a large scale dataset with billions of samples is used to learn DNNs for pairwise semantic ranking. We randomly split out 20\% samples from the training data to form the validation set.

\paragraph{Methods and Settings.} In \cite{HashNet_ICML2015}, the authors compared HashedNets against several DNN compression approaches, and showed HashedNets performs consistently the best, including the low-rank decomposition \cite{RedundDNN_NIPS2013}. Under the same settings, we compare \emph{FunHashNN} with \emph{HashedNets}\footnote{HashedNets code downloaded from http://www.cse.wustl.edu/$\sim$wenlinchen/project/HashedNets/index.html}  and a standard neural network without compression. All activation functions are chosen as ReLU. 

The settings of FunHashNN are tested in two scenarios. First, we will fix to use FunHashNN in Figure~\ref{fig:model}(b) without extensions, and then compare the effects of compression by FunHashNN and HashedNets. Second, we compare different configurations of FunHashNN itself, including the number $U$ of seeds, the layer of reconstruction network $g(\cdot)$, and extension with the dual space hashing. Hidden layers within $g(\cdot)$ keep using tanh as activation functions. Results by the multi-hops extension of FunHashNN will be included in another ongoing paper for systematic comparisons.

\subsection{Varying Compression Ratio}
\vspace{-2mm}
To test robustness, we vary the compression ratio with (1) a fixed virtual network size (i.e., the size of $\mb{V}^{\ell}$ in each layer), and then with (2) a fixed memory size (i.e., the size of $\mb{w}^{\ell}$ in each layer). Three-layer (1 hidden layer) and five-layer (3 hidden layers) networks are investigated. In experiments, we vary the compression ratio geometrically within $\{1, \frac{1}{2}, \frac{1}{4}, \ldots, \frac{1}{64}\}$. For FunHashNN, this comparison sticked to use 4 hash functions, 3-layer $g(\cdot)$, and without dual space hashing.

\begin{figure}[!h]
 \centering
 \begin{minipage}{\textwidth}
 \centering
  \begin{minipage}{.45\textwidth}
  \centering
   \includegraphics[width=\textwidth]{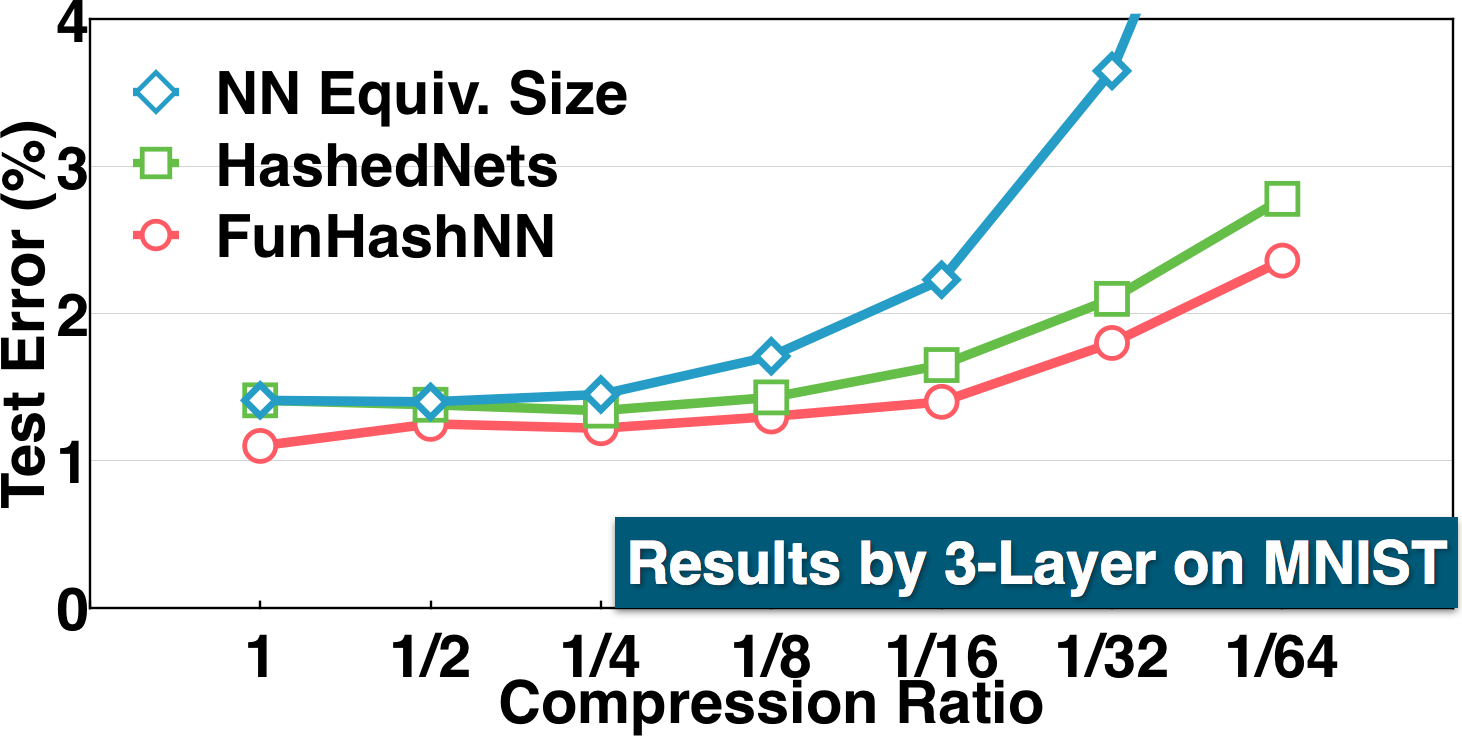}
   \end{minipage}
   \hspace{5mm}
   \begin{minipage}{.45\textwidth}
  \centering
   \includegraphics[width=\textwidth]{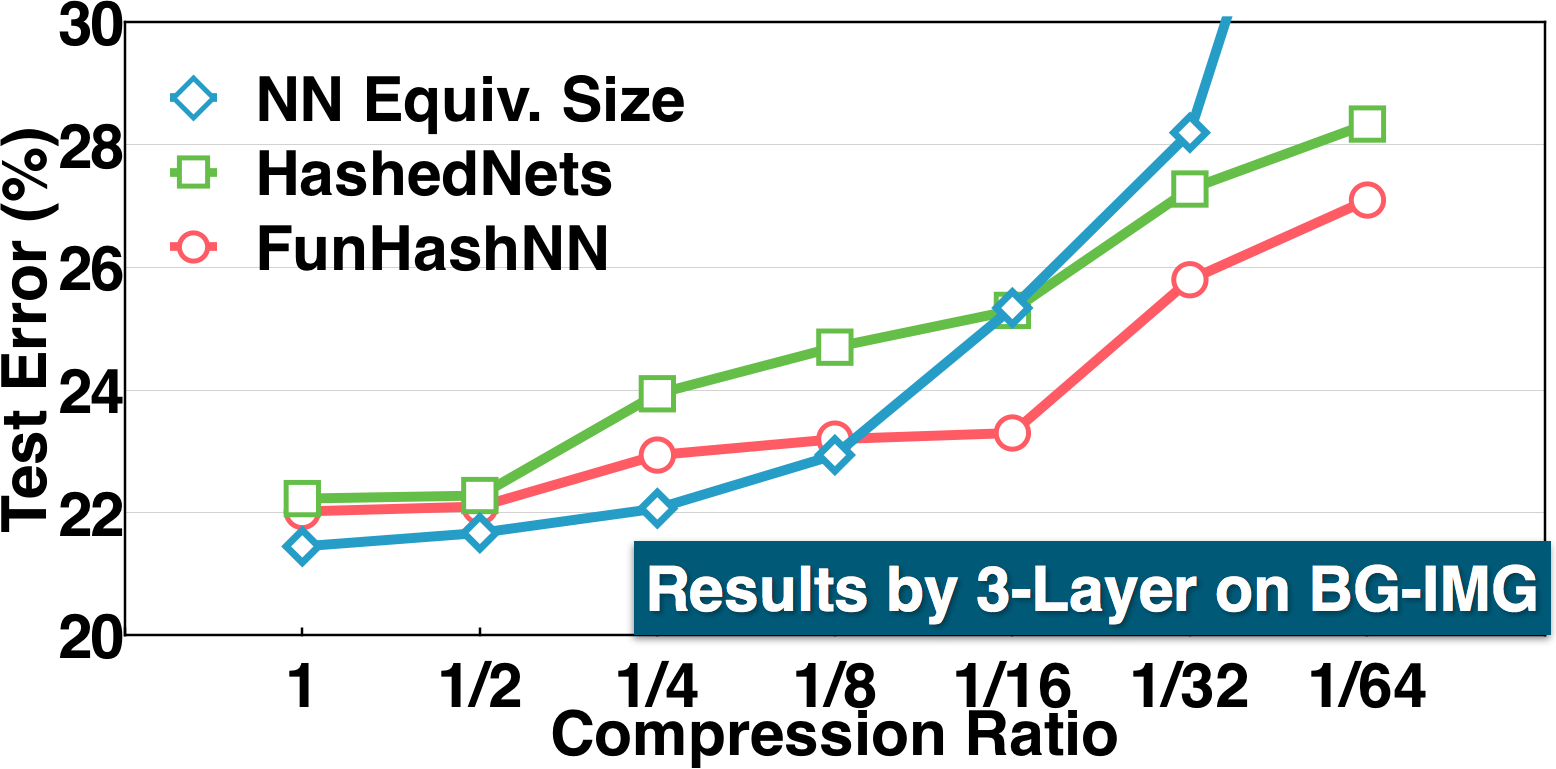}
   \end{minipage}
   \end{minipage}\\ \vspace{3mm}
   \begin{minipage}{\textwidth}
   \centering
   \begin{minipage}{.45\textwidth}
  \centering
   \includegraphics[width=\textwidth]{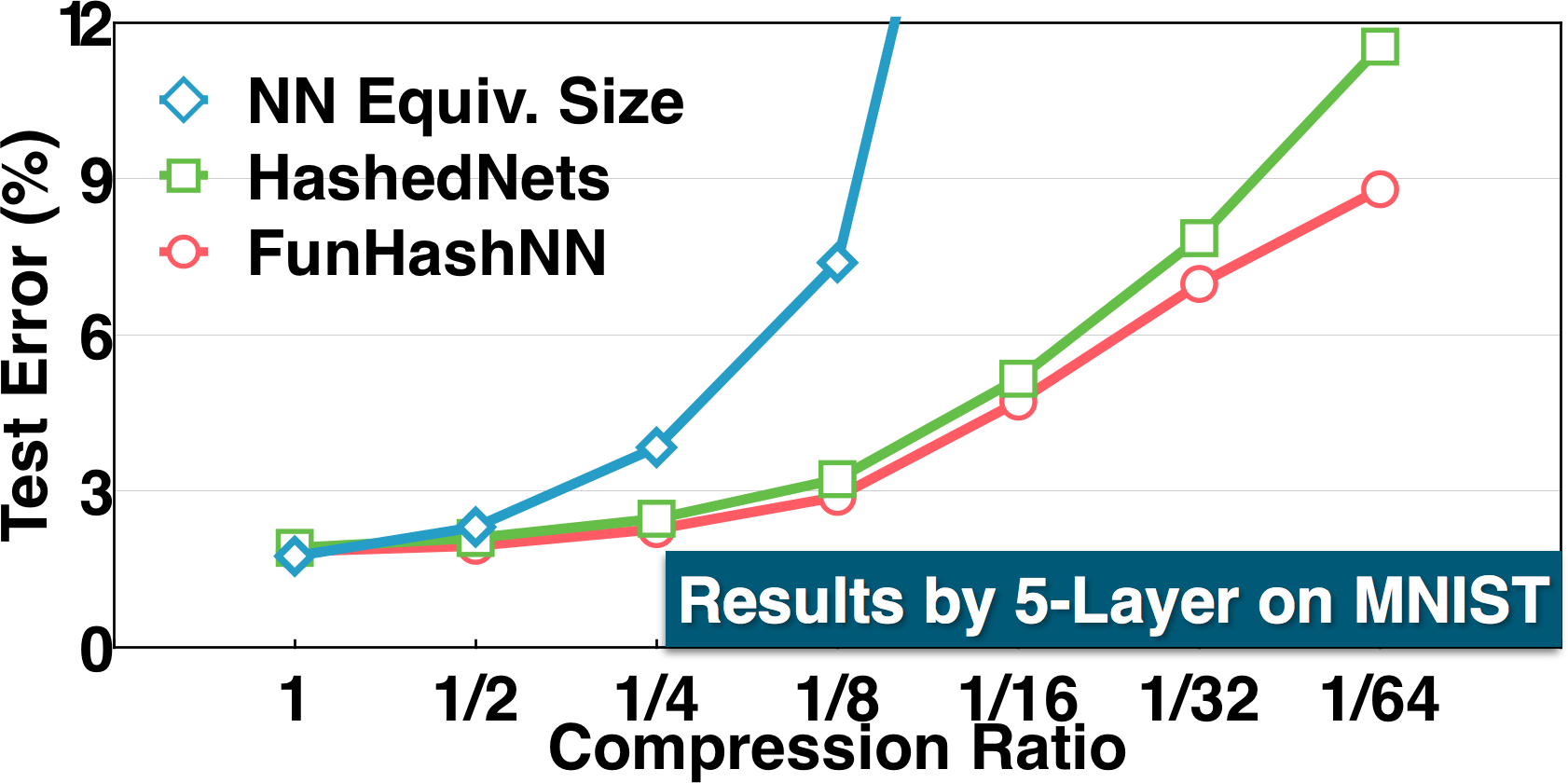}
   \end{minipage}
   \hspace{5mm}
   \begin{minipage}{.45\textwidth}
  \centering
   \includegraphics[width=\textwidth]{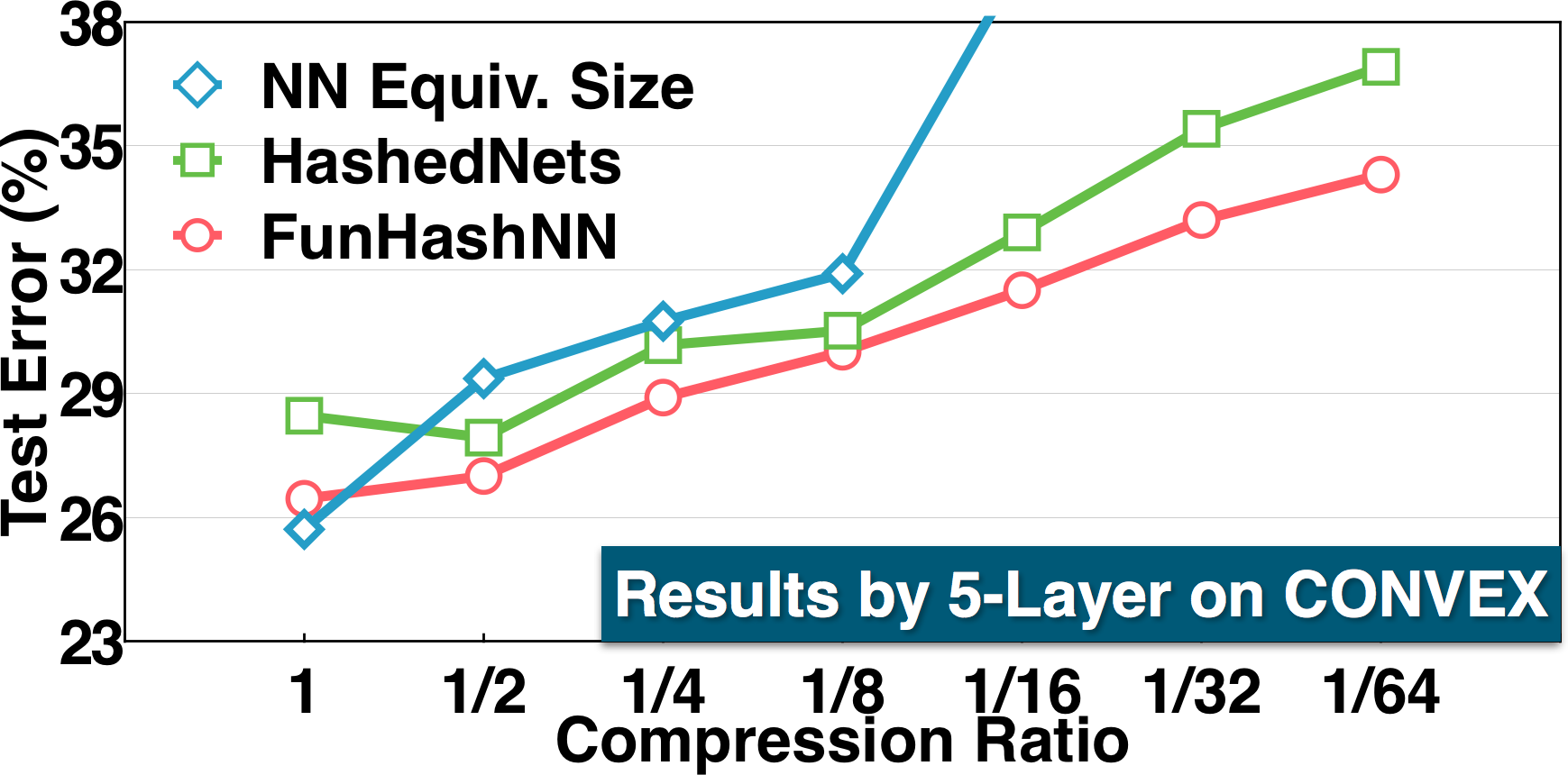}
   \end{minipage}
   \end{minipage}
   \vspace{-3mm}
  \caption{Testing errors by varying compression ratio with a fixed virtual network size.}\label{fig:res_varycomp_fixnet}
\end{figure}

\paragraph{With Virtual Network Size Fixed.} The hidden layer for 3-layer nets initializes at 1000 units, and for 5-layer nets starts at 100 units per layer. As the compression ratio ranges from 1 to 1/64 with a fixed virtual network size, the  memory decreases and it becomes increasingly difficult to preserve the classification accuracy. The testing errors are shown in Figure~\ref{fig:res_varycomp_fixnet}, where standard neural networks with equivalent parameter sizes are included in comparison. FunHashNN shows robustly effective compression against the compression ratios, and persistently produces better prediction accuracy than HashedNets. It should be noted, even when the compression ratio equals to one, FunHashNN with the reconstruction network structure is still not equivalent to HashedNets and performs better.

\begin{figure}[!h]
 \centering
 \begin{minipage}{\textwidth}
 \centering
  \begin{minipage}{.45\textwidth}
  \centering
  \includegraphics[width=\textwidth]{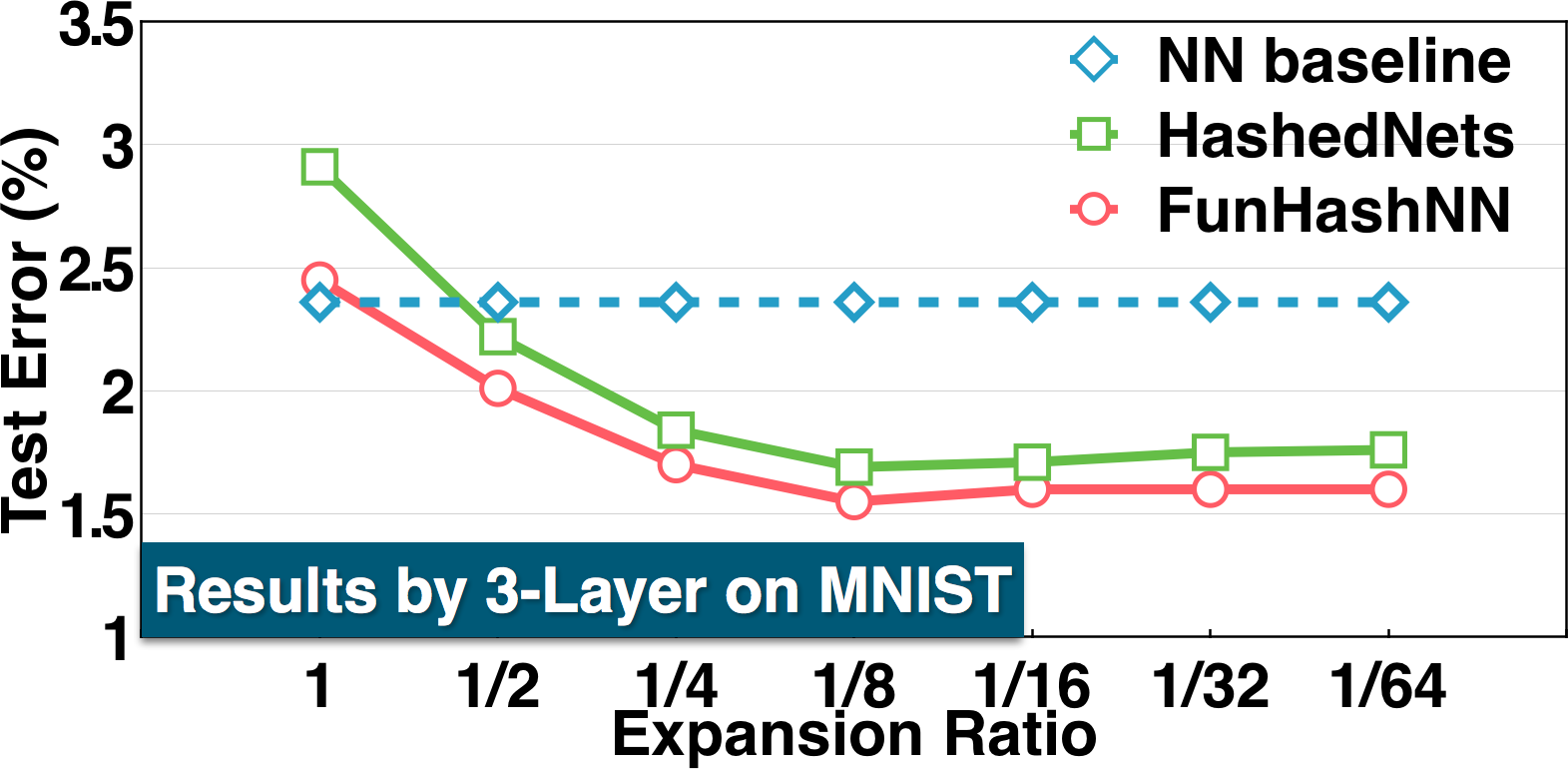}
  \end{minipage}
  \hspace{5mm}
  \begin{minipage}{.45\textwidth}
  \centering
  \includegraphics[width=\textwidth]{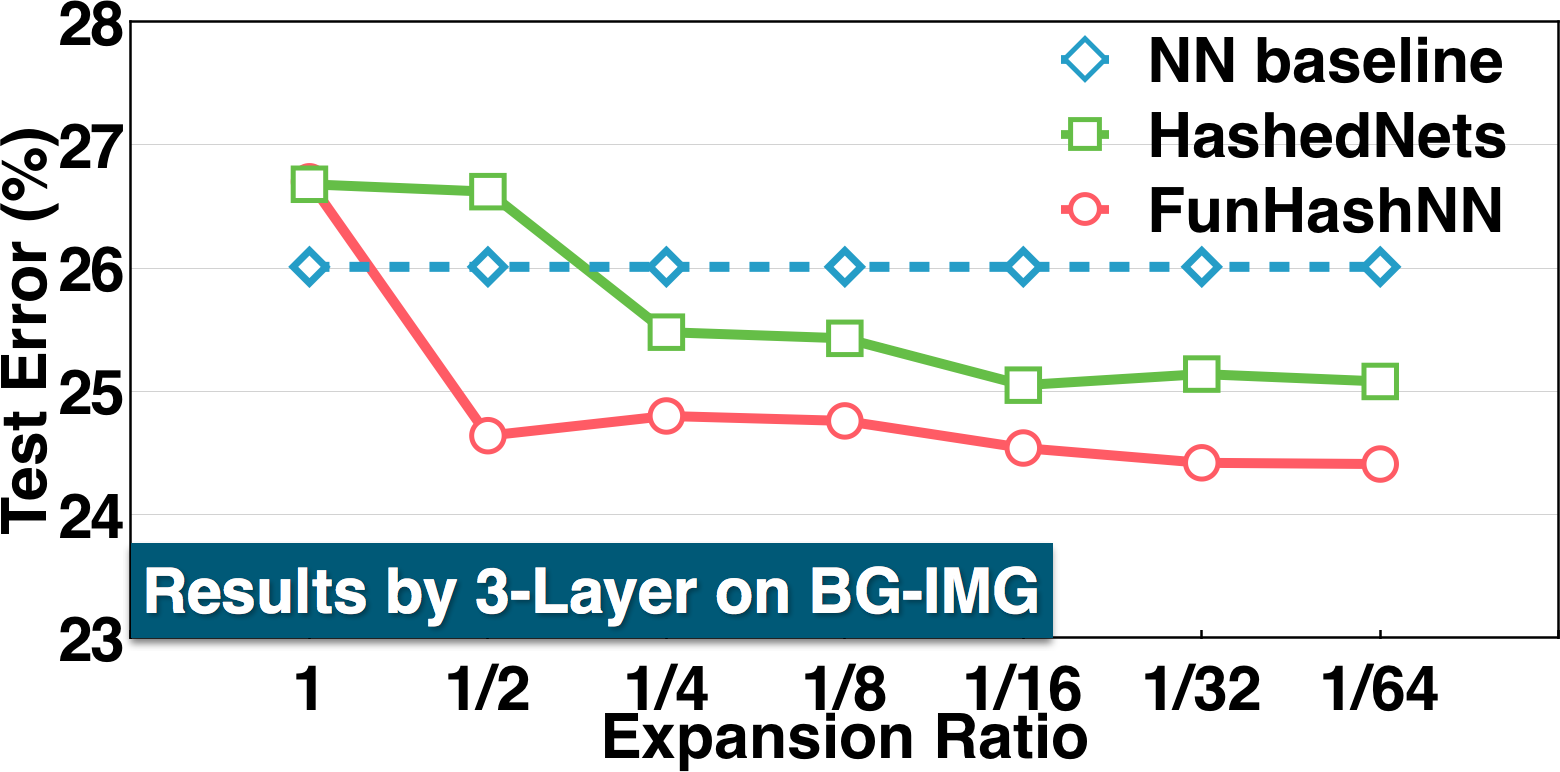}
  \end{minipage}
  \end{minipage}\\ \vspace{3mm}
  \begin{minipage}{\textwidth}
  \centering
  \begin{minipage}{.45\textwidth}
  \centering
  \includegraphics[width=\textwidth]{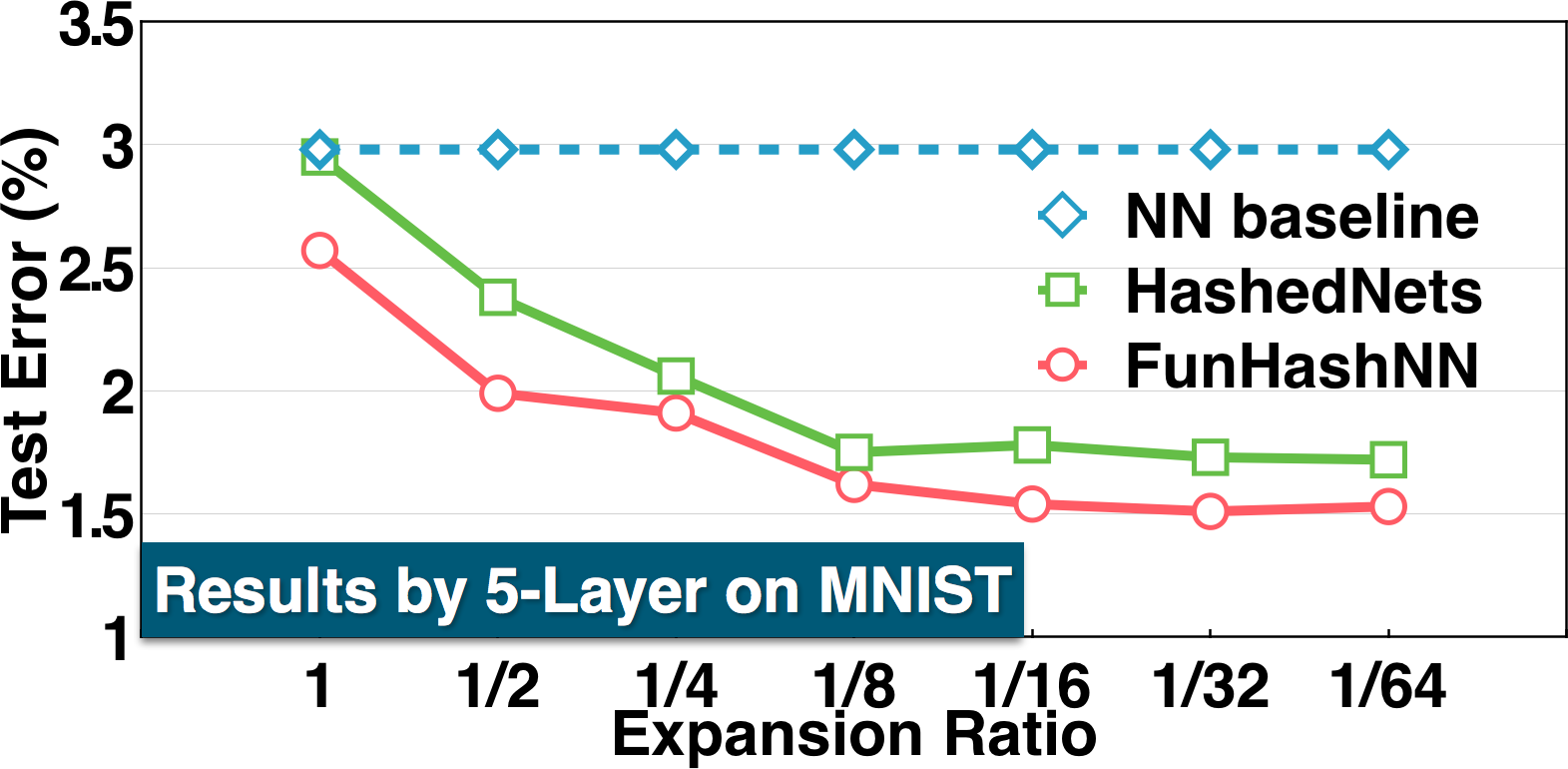}
  \end{minipage}
  \hspace{5mm}
  \begin{minipage}{.45\textwidth}
  \centering
  \includegraphics[width=\textwidth]{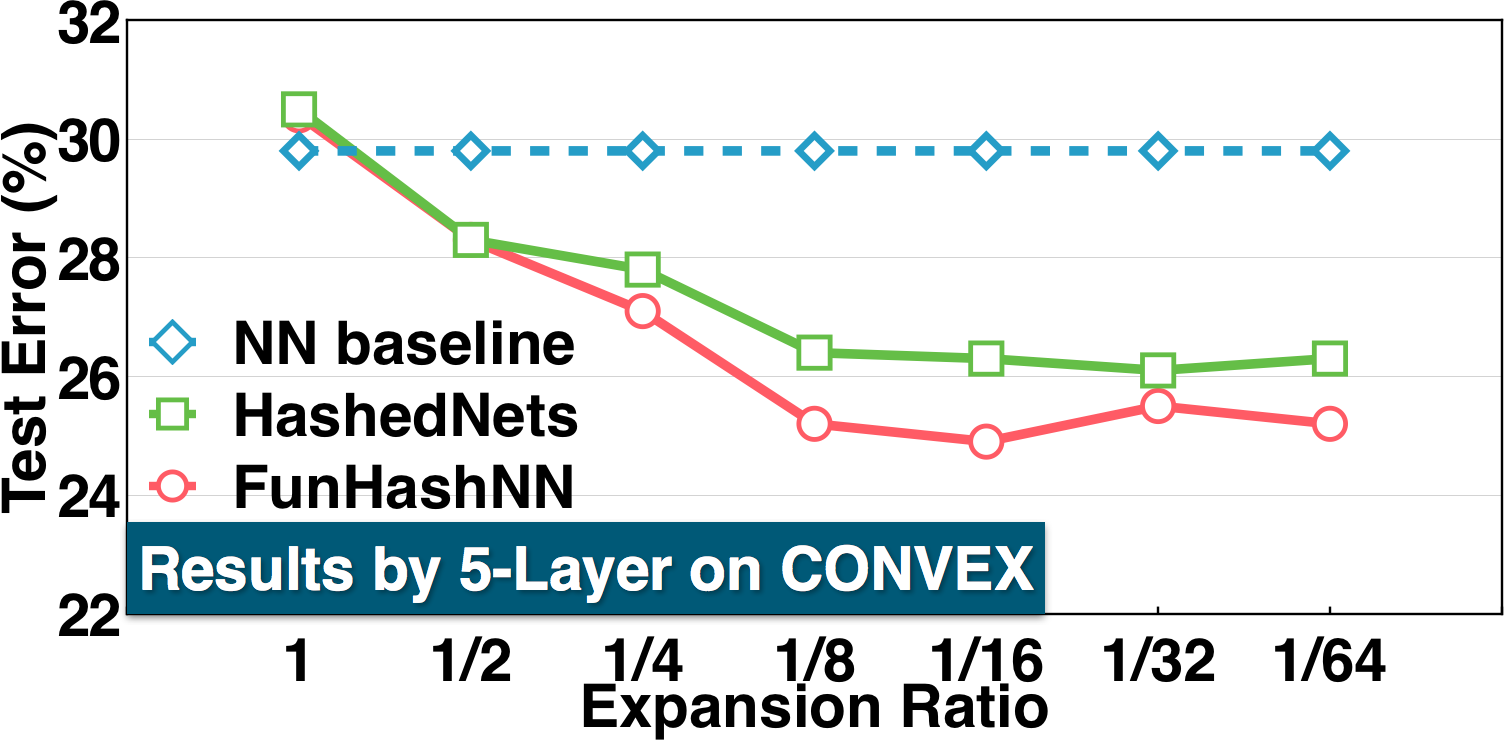}
  \end{minipage}
  \end{minipage}
  \vspace{-3mm}
  \caption{Testing errors by varying compression (expansion) ratio with a fixed memory storage.}\label{fig:res_varycomp_fixmem}
\end{figure}

\paragraph{With Memory Storage Fixed.} We change to vary the compression ratio from 1 to 1/64 with a fixed memory storage size, i.e., the size of the virtual network increases while the number of free parameters remains unchanged. In this sense, we'd better call it expansion instead of compression. Both 3-layer and 5-layer nets initialize at 50 units per hidden layer. The testing errors in this scenario are shown in Figure~\ref{fig:res_varycomp_fixmem}. At all compression (expansion) ratios on each dataset, FunHashNN performs better than or at least comparably well compared to HashedNets.

\subsection{Varying Configurations of FunHashNN}
\vspace{-2mm}
On 3-layer nets with compression ratio $1/8$, we vary the configuration dimensions of FunHashNN, including the number of hash functions ($U$), the structure of layers of the reconstruction network $g(\cdot)$, and whether dual space hashing is turned on. Since it is impossible to enumerate all probable choices, $U$ is restricted to vary in $\{2,4,8,16\}$. The structure of $g(\cdot)$ is chosen from $2\sim4$ layers, with $U\times1$, $U\times0.5U\times1$, $U\times U\times0.5U\times1$ layerwise widths, respectively. We denote U$x$-G$y$ as $x$ hash functions and $y$ layers of $g(\cdot)$, and a suffix -D indicates the dual space hashing. 

Table~\ref{tab:res_varyconfig} shows the performances of FunHashNN with different configurations on \texttt{MNIST}. The observations are summarized below. First, the series from index (0) to (1.x) fixes a 3-layer $g(\cdot)$ and varies the number of hash functions. As listed, more hash functions do not ensure a better accuracy, and instead U4-G3 performs the best, perhaps because too many hash functions potentially brings too many partial collisions. Second, the series from (0) to (2.x) fixes the number of hash functions and varies the layer number in $g(\cdot)$, where three layers performs the best mainly due to its strongest representability. Third, indices (3.x) show further improved accuracies using dual space hashing.

\begin{minipage}{\textwidth}
 \centering
 \begin{minipage}{.4\textwidth}
 \centering
  \makeatletter\def\@captype{table}\makeatother
   \caption{Performances on \texttt{MNIST} by various configurations of FunHashNN.}\label{tab:res_varyconfig}
    \begin{tabular}{llc}
    \toprule
    Index & Config     & Test Error(\%)  \\
    \midrule
    (0)  & U4-G3 &  1.32 \\
    (1.1) & U2-G3 & 1.42 \\
    (1.2) & U8-G3 & 1.39 \\
    (1.3) & U16-G3 & 1.40 \\
    (2.1) & U4-G2 & 1.34 \\
    (2.2) & U4-G3 & 1.28 \\
    (3.1) & U2-G3-D & 1.36  \\
    (3.2) & U4-G3-D & 1.24 \\
    (3.3) & U8-G3-D & 1.27\\
    \bottomrule
  \end{tabular}
  \end{minipage}\hspace{8mm}
  \begin{minipage}{.52\textwidth}
     \centering \vspace{3mm}
     \includegraphics[width=\textwidth]{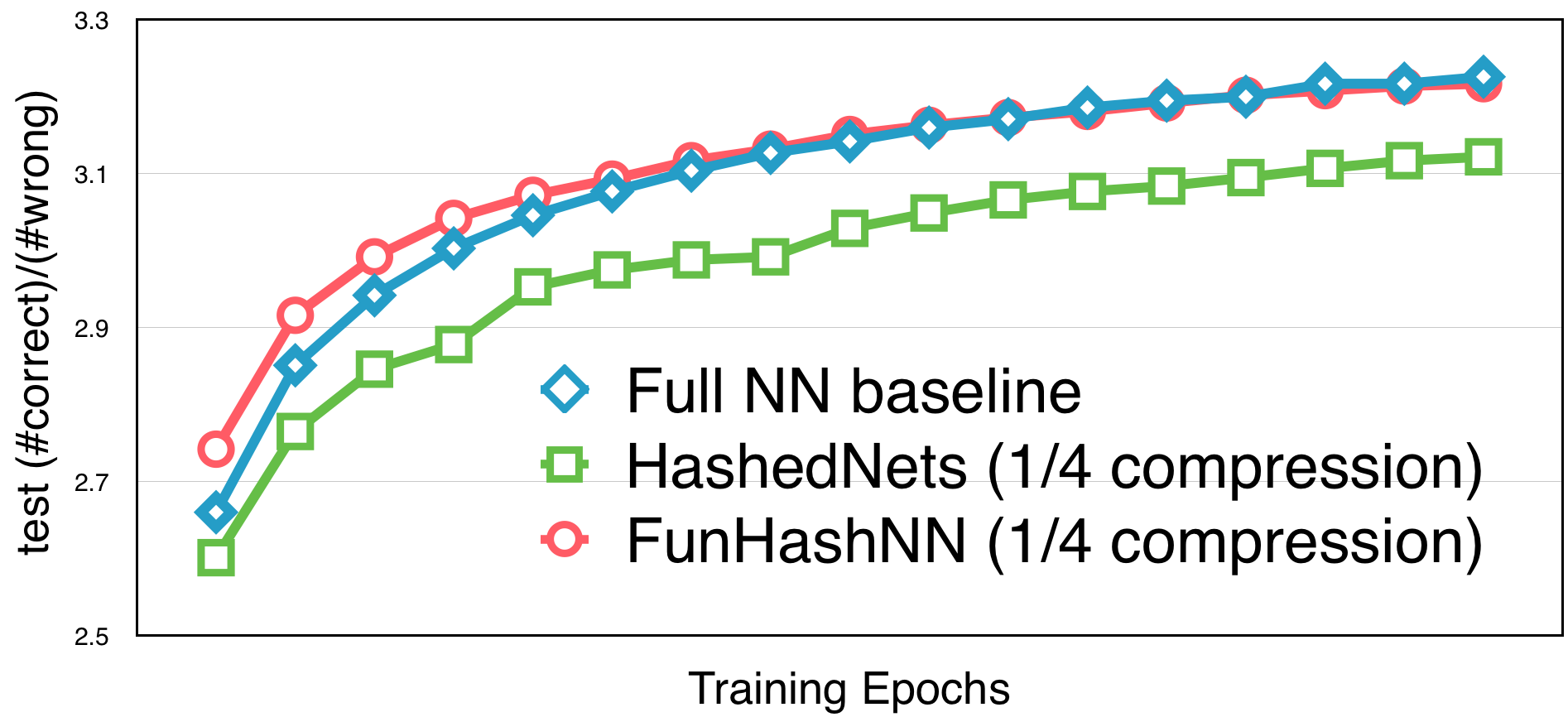}\\ \vspace{-3mm}
     \makeatletter\def\@captype{figure}\makeatother
     \caption{Performances for pairwise semantic ranking. Testing correct-to-wrong pairwise ranking ratios (the larger the better) are plotted versus the number of training epochs.} \label{fig:res_hash_ltr}
  \end{minipage}
\end{minipage}

\subsection{Pairwise Semantic Ranking}
\vspace{-2mm}
Finally, we evaluate the performance of FunHashNN on semantic learning-to-rank DNNs. The data is collected from logs of a commercial search engine, with per clicked query-url being a positive sample and per non-clicked being a negative sample. There are totally around 45B samples. We adopt a deep convolutional structured semantic model similar to \cite{DSSM_CIKM2013,DSSM_CIKM2014}, which is of a siamese structure to describe the semantic similarity between a query and a url title. The network is trained to optimize the cross entropy for each pair of positive and negative samples per query.

The performance is evaluated by correct-to-wrong pairwise ranking ratio on testing set. In Figure~\ref{fig:res_hash_ltr}, we plot the performance by a baseline network as training proceeds, compared to FunHashNN and HashNet both with 1/4 compression ratio. With $U=4$ hash functions, FunHashNN performs better than HashedNets throughout the training epochs, and even comparable to the full network baseline which requires 4 times of memory storage. The deterioration of HashedNets probably comes from many inappropriate collisions on word embeddings, especially for words of high frequencies.

\section{Conclusion and Future Work}\label{sec:conclusion}
\vspace{-3mm}
This paper presents a novel approach FunHashNN for neural network compression. Briefly, after adopting multiple low-cost hash functions to fetch values in compression space, FunHashNN employs a small reconstruction network to recover each entry in an matrix of the original network. The reconstruction network is plugged into the whole network and learned jointly.  The recently proposed HashedNets \cite{HashNet_ICML2015} is shown as a degenerated special case of FunHashNN. Extensions of FunHashNN with dual space hashing and multi-hops are also discussed. On several datasets, FunHashNN demonstrates promisingly high compression ratios with little loss on prediction accuracy.

As future work, we plan to further systematically analyze the properties and bounds of FunHashNN and its extensions. More industrial applications are also expected, especially on mobile devices. This paper focuses on the fully-connected layer in DNNs, and the compression performance on other structures (such as convolutional layers) is also planned to be studied.  As a simple and effective approach, FunHashNN is expected to be a standard tool for DNN compression.



\newpage
\protect\nocite{*}
\medskip
\small
\bibliographystyle{abbrv}

\end{document}